\documentclass[10pt]{article} 
\usepackage[preprint]{tmlr}
\newif\iftmlrcamera
\tmlrcamerafalse  
\tmlrcameratrue

\newcommand{\repoURL}{%
    \iftmlrcamera
      https://github.com/semueller/stobb\_cobot
    \else
      https://anonymous.4open.science/r/cobot\_tmlr-F583/README.md
    \fi
}

\newcommand{\repoName}{%
    \iftmlrcamera
      github.com/semueller/stobb\_cobot
    \else
      anonymous.4open.science/r/cobot\_tmlr-F583
    \fi
}
\usepackage{underscore}

\usepackage{amsmath,amsfonts,bm}

\newcommand{\sobs}{\mathcal{B}}
\newcommand{\boxsystem}{B}
\newcommand{\itemset}{I}

\newcommand{\blbox}{f} 

\newcommand{\sample}{x}
\newcommand{\inputdomain}{\R^d}
\newcommand{\observation}{o}
\newcommand{\lobservation}{\mathcal{O}}
\newcommand{\bobox}{b} 

\newcommand{\localexplainer}{\Phi}
\newcommand{\binarization}{\mathcal{I}}
\newcommand{\bink}{\binarization^{k\uparrow}}

\newcommand{\binthree}{\binarization^{3\uparrow}}

\newcommand{\outputspace}{\text{C}}

\def\eqref#1{equation~\ref{#1}}

\def\1{\bm{1}}

\DeclareMathAlphabet{\mathsfit}{\encodingdefault}{\sfdefault}{m}{sl}
\SetMathAlphabet{\mathsfit}{bold}{\encodingdefault}{\sfdefault}{bx}{n}

\newcommand{\R}{\mathbb{R}}

\usepackage{placeins}

\usepackage{url}

\usepackage{multicol}
\usepackage{enumitem}

\usepackage{booktabs,tabularx,ragged2e,makecell,nicematrix}
\newcolumntype{Y}{>{\RaggedRight\arraybackslash}X}

\usepackage{tikz}
\usetikzlibrary{external}
\usetikzlibrary{arrows.meta,positioning}
\usepackage{pgfplots} 
\pgfplotsset{compat=1.18}

\usepackage{amssymb}

\usepackage{natbib}

\newcommand{\exbibitem}[1]{\item}

\usepackage{makecell}
\usepackage{graphicx}
\usepackage{booktabs}
\usepackage{longtable}
\usepackage{adjustbox} 
\usepackage{subfig}
\usepackage{subcaption}

\usepackage{float} 
\usepackage{pifont}
\usepackage{xcolor}
\usepackage{multirow}
\usepackage{rotating}
\usepackage{booktabs}
\usepackage{bm}

\usepackage[normalem]{ulem}

\usepackage{makecell}

\usepackage{algorithm}
\usepackage[noend]{algpseudocode}
\usepackage{caption}
\usepackage{subcaption}
\usepackage{float}
\usepackage{etoolbox} 

\usepackage[colorlinks=true,allcolors=blue]{hyperref}
\usepackage[nameinlink]{cleveref}
\usepackage[german,british]{babel} 

\usepackage{listings}
\usepackage[font=small,labelfont=bf]{caption}
\usepackage{subcaption} 

\usepackage[misc]{ifsym} 
\usepackage{makecell}

\definecolor{mutedorange}{RGB}{230, 120, 60} 
\definecolor{mutedblue}{RGB}{90, 142, 185}
\definecolor{mutedgreen}{RGB}{100, 150, 100}
\definecolor{mutedpurple}{RGB}{178, 132, 190}

\newcommand{\ie}{i.e.,}
\newcommand{\eg}{e.g.,}
\newcommand{\vF}{van Fraassen}
\newcommand{\bb}{black-box}

\newcommand{\cobot}{CoBoT}
\newcommand{\cobotlong}{{Constructive Box Theoriser}}

\newcommand{\stob}{SToBB}

\title{Scientific Theory of a Black-Box: A Life Cycle-Scale\\ XAI Framework Based on Constructive Empiricism}  

\author{\name Sebastian M\"uller \email semueller@uni-bonn.de \\
  \addr MLAI Group, University of Bonn\\
  Lamarr Institiute, Bonn
  \AND
  \name Vanessa Toborek \\
  \addr MLAI Group, University of Bonn\\
  Lamarr Institiute, Bonn
  \AND
  \name Eike Stadtl\"ander \\
  \addr MLAI Group, University of Bonn\\
  Lamarr Institiute, Bonn
  \AND
  \name Tam\'as Horv\'ath \\
  \addr MLAI Group, University of Bonn\\
  Lamarr Institiute, Bonn\\
  Fraunhofer IAIS, Sankt Augustin
  \AND
  \name Brendan Balcerak Jackson \\
  \addr Lamarr Institiute, Bonn\\
  University of Bonn
  \AND
  \name Christian Bauckhage \\
  \addr MLAI Group, University of Bonn\\
  Lamarr Institiute, Bonn\\
  Fraunhofer IAIS, Sankt Augustin}

\def\month{MM} 
\def\year{YYYY} 
\def\openreview{\url{https://openreview.net/forum?id=XXXX}} 

\begin{document}
%


%


%

%
\maketitle    
 \begin{abstract}

Explainable AI (XAI) offers a growing number of algorithms that aim to answer specific questions about black-box models.
What is missing is a principled way to consolidate explanatory information about a fixed black-box model into a persistent, auditable artefact, that accompanies the \bb{} throughout its life cycle. 
We address this gap by introducing the notion of a \textit{scientific theory of a black box} (\stob{}). 
Grounded in Constructive Empiricism, a \stob{} fulfils three obligations: 
(i) \textit{empirical adequacy} with respect to all available observations of black-box behaviour, (ii) \textit{adaptability} via explicit update commitments that restore adequacy when new observations arrive, and (iii) \textit{auditability} through transparent documentation of assumptions, construction choices, and update behaviour.
We operationalise these obligations as a general framework that specifies an extensible observation base, a traceable hypothesis class, algorithmic components for construction and revision, and documentation sufficient for third-party assessment. Explanations for concrete stakeholder needs are then obtained by querying the maintained record through interfaces, rather than by producing isolated method outputs.
As a proof of concept, we instantiate a complete \stob{} for a neural-network classifier on a tabular task and introduce the \cobotlong{} (\cobot{}) algorithm, an online procedure that constructs and maintains an empirically adequate rule-based surrogate as observations accumulate. 
Together, these contributions position \stob{}s as a life cycle-scale, inspectable point of reference that supports consistent, reusable analyses and systematic external scrutiny.
\end{abstract}

\section{Introduction}
\label{sec:intro}

Explainable AI (XAI) provides methods for analysing \bb{} decisions. Each method builds on its own assumptions and evaluation criteria and is intended to answer a specific question for specific users \citep{speith2022taxonomies,tomsett2018stakeholder,nauta2023anecdotal,mersha2024surveyxai}.
While XAI offers methods for the individual needs, these methods are typically applied in an ad hoc manner, producing isolated outputs that may need to be recomputed for similar questions and that can yield incompatible characterisations \citep{krishna2022disagreement} of the same model.
To address this problem, we suggest a perspective that treats explanatory information as an evolving artefact maintained across the \bb{}'s life cycle, rather than as a sequence of disconnected method outputs.
Accordingly, such a foundation should aim to ensure adequacy, adaptability, and auditability over time, providing a stable basis for consistent analyses while remaining open to new forms of inquiry.
We argue that one way forward is to approach explainability for machine learning systems in the same role that scientific theories play in science: designed not to be answers to any particular question outright, but to be persistent and auditable consolidations of knowledge that provide a shared point of reference and support diverse forms of inquiry.
To this end, we propose and develop the idea of a \textit{scientific theory of a \bb{}} (\stob{}).

To consider a more concrete scenario, one can imagine a fixed classifier trained on the Abalone dataset \citep{abalone_1} that predicts whether an individual abalone has reached a target age, relevant to harvesting decisions and pricing. During development, model developers probe the classifier globally to understand which input features (\eg{} physical measurements of shell size or weight) influence predictions and where the model fails, using global summaries and debugging-oriented explanations. After deployment, operators and employees in an aquaculture setting may request local or contrastive explanations for specific batches, \eg{} why a given lot is classified as not ready for harvest and what changes in measurements would flip that assessment. At the same time, sustainability officers or external auditors may periodically inspect the model's behaviour over time, asking whether different abalone subpopulations are treated consistently and whether model decisions reflect sustainability obligations. 
Wholesalers seeking to buy abalone may request justification for the asked price, which could be supported by explanations that demonstrate the relevant qualities of the batch, such as predicted age, size, and compliance with sustainability constraints.
All of these inquiries concern the same underlying classifier and lead to similar probings of it, but arise in different contexts, \ie{} different stages of the system’s life cycle and from different stakeholders who vary in intent and background knowledge.
To address a diverse set of inquiries, a \stob{} organises shared explanatory information as a maintained record that can accumulate over time and be re-used in different contexts; concrete, context-appropriate explanations are then obtained by querying this record.

For the conceptual grounding of the \stob{} we turn to philosophy of science, which offers established accounts of scientific theories and their function. In particular, \textit{Constructive Empiricism} (CE) \citep{bvf1980constructiveempiricism} aligns with two central requirements in XAI: fidelity to observed model behaviour and pragmatic usefulness to different users. In CE, scientific theories are judged by their \textit{empirical adequacy}, that is, their agreement with all observed phenomena, and by their \textit{usefulness} to the scientist. Acceptance of a theory rests on two commitments: conviction in the theory’s empirical adequacy and a willingness to use and adapt it as new observations arise. A theory is therefore part of a continuous process in which observations either reaffirm its adequacy or demand refinement. Notably, CE does not assume that theories describe an underlying truth that lies beyond what is observable. This stance avoids overclaiming access to the supposed hidden ``true reasoning'' of a black-box while still providing a structured and auditable basis for analysis.

To operationalise the concept of a \stob{} for XAI, \textit{surrogate models} offer a natural starting point. Surrogates are inherently interpretable models designed to approximate the \bb{}’s input-output mapping \citep{bastani2017extraction,molnar2020interpretable}. They are typically applied post hoc and independently of the \bb{}’s architecture \citep{speith2022taxonomies} and may approximate behaviour locally \citep{ribeiro2016lime,lundberg2017unified,ribeiro2018anchors} or globally \citep{bastani2017extraction}. 
In contrast to theories in CE, they are not generally required to achieve perfect agreement with the \bb{} on all observed data, nor are they systematically adapted when inconsistencies appear. This limits their epistemic value over time.
Constructive Empiricism provides a principled basis to recast surrogates as scientific theories: to achieve empirical adequacy, the surrogate must agree with the \bb{} on all available observations, and to justify acceptance demands adaptability to preserve adequacy as new observations accumulate.

Scientific theories are also documented, scrutinised, and evaluated within a community. 
The same principle motivates a \stob{}, which must make explicit its assumptions, construction choices, and update behaviour to enable third parties (\eg{} auditors or operators) to independently assess its merit, scope, and limitations.
Detailed transparency requirements and their connection to existing documentation practices are discussed in \Cref{sec:context}.

It is important to emphasise that we consider the \stob{} a complementary perspective in XAI and not a universal solution. It addresses a specific gap: the lack of a persistent, expanding, auditable representation of explanatory information that accompanies a \bb{} as observations accumulate. As a result, \stob{}s serve as a life cycle-scale, inspectable point of reference for explanation: empirical adequacy provides a checkable baseline, while decoupling explanatory information from its context-dependent presentation enables consistent, reusable analyses across stakeholders.
Throughout this paper, we treat the black-box model as fixed, meaning that its decision behaviour does not change while explanatory information is collected. We return to this assumption and its implications in the discussion.

In this paper we make the following contributions:
\begin{itemize}
    \item We introduce the notion of a \stob{}, a documented and evolving artefact that represents a fixed black-box model via an interpretable surrogate, an observation base, and explicit update and documentation commitments.
    
    \item As a conceptual prerequisite to this definition, we operationalise CE for XAI by translating empirical adequacy, acceptance as commitment, and pragmatic virtues into concrete design and documentation obligations for constructing and maintaining \stob{}s.
    
    \item We present a proof-of-concept that instantiates a complete \stob{} for a neural-network classifier on a tabular task; to this end, we introduce the \textit{\cobotlong{}} (\cobot{}) algorithm as a generic procedure for tabular domains, which may be of independent interest.
\end{itemize}


Outline: \Cref{sec:context} motivates the structural gap in the current XAI landscape that a \stob{} aims to address. Subsequently, \Cref{sec:CEXAI} introduces the central notions of Constructive Empiricism and 
transfers the concepts to XAI. This yields an overview of the properties a \stob{} must fulfill. Based on this, \Cref{sec:def} defines the constituents of the \stob{} and discusses its properties. \Cref{sec:proc_poc} goes on to describe and exemplify the process to create a \stob{}, also introducing the \cobot{} algorithm. \Cref{sec:discussion} provides the discussion and conclusion.

\section{Context}
\label{sec:context}

In this paper we argue for a perspective that treats explanatory information regarding a \bb{} as something that is collected and integrated over time.
In this section we examine practical motivations for this perspective, focusing on two aspects: 
(i) heterogeneous stakeholder needs, and (ii) transparency needs that arise throughout the AI system life cycle.

\paragraph{(i) Stakeholders and explanatory needs}
The XAI literature recognizes a diverse set of stakeholders who require insight into the behaviour of AI systems. These groups include, but are not limited to, developers, regulatory entities, operators, domain experts, and affected users \citep{tomsett2018stakeholder,arrieta2020xai,langer2021stakeholder,mersha2024surveyxai}. 
\citet{langer2021stakeholder} emphasize that individuals may belong to several stakeholder groups and that individuals within a group can differ in expertise, background knowledge, and expectations. 
Research on user needs further shows that these stakeholders ask different kinds of explanatory questions and rely on different levels of abstraction \citep{liao2020questionbank,liao2021xaiux}.

\paragraph{(ii) Transparency needs across the AI life cycle}
Stakeholder groups largely identical to those identified in the XAI literature also appear in several Trustworthy ML frameworks and standards.\footnote{IEEE~P7001 \citep{ieee2021std7001}, ISO/IEC~42001 \citep{ISOIEC42001_2023}, the OECD AI Recommendation \citep{oecd2024trustworthy}, the NIST AI Risk Management Framework \citep{nist2023airmf}, the Fraunhofer Guidelines for Trustworthy AI \citep{poretschkin2023guideline}, and the EU Ethics Guidelines for Trustworthy AI \citep{euhleg2019ethics} that now underpin the EU AI Act \citep{EU2024AIAct}} 
All frameworks embed AI systems within a broader \textit{life cycle} that spans design, development, testing, verification, deployment, and monitoring. Although the frameworks differ in focus -- from risk management to organisational governance -- they identify similar qualities for trustworthiness, such as accountability, human agency, technical robustness, compliance, and transparency. Importantly, transparency is regarded not only as a property in itself but also as an enabler for these other qualities, and transparency needs arise \textit{continuously} at all stages of the life cycle \citep{nist2023airmf,euhleg2019ethics,oecd2024trustworthy}.
Explainability plays an important role in addressing some of these transparency needs. 
As the Abalone example in the introduction motivates: Developers may rely on debugging-oriented analyses early in the life cycle, end-users may require local or contrastive decision explanations to verify fair treatment or obtain actionable information, and operators or auditors may need system-level evidence over time to assess compliance and orderly operation.

\paragraph{The structural gap in the XAI landscape}
While XAI acknowledges diverse stakeholders, the life cycle perspective is largely absent. Surveys of XAI methods \citep{arrieta2020xai,speith2022taxonomies,mersha2024surveyxai} show that research predominantly develops techniques for isolated explanatory tasks. Although these methods see successful practical use \citep{liao2021xaiux,arya2019onefitall}, they remain compartmentalised even when probing the \textit{same} black-box model. In particular, explanations are produced independently of each other, which may lead to incompatible descriptions of model behaviour, a phenomenon known as the \textit{disagreement problem} \citep{krishna2022disagreement}, or to redundant computations over time.

Recall that stakeholders may have different explanatory goals and differ in their background knowledge. Despite these differences, all inquiries concern the same underlying model. If explanatory information is collected only for the immediate question at hand, it remains siloed and must be recomputed when similar questions arise later in the life cycle.
A structured approach that records and consolidates explanatory information therefore enhances efficiency, consistency, and reuse across stakeholders. We propose to treat explanatory information not as a set of disconnected method outputs, but as a persistent \textit{record} that aggregates observations and remains consistent as it evolves. Such a record is not itself an explanation, but a representational foundation from which diverse explanations can be derived. 
To the best of our knowledge, there is no artefact in the current XAI landscape intended to simultaneously (i) record, (ii) represent, and (iii) continuously update explanatory information for a fixed \bb{} as observations accumulate, while remaining auditable and reusable across the system life cycle. The \stob{} framework fills this gap by treating explanatory information as a maintained record rather than a sequence of disconnected method outputs.

What would such an artefact be? From the perspective of the philosophy of science, something that represents the behaviour of a system and serves as a common foundation from which diverse explanations can be derived is a \textit{scientific theory}. Following this perspective we turn, in particular, toward the operationalisation of CE \citep{bvf1980constructiveempiricism} as a framework for structuring such a record. The semantic conception of theories in CE and the emphasis on empirical adequacy and updateability align closely with the requirements identified above. The next section introduces the core notions of CE and discusses their translation to XAI.

\paragraph{Documentation as a second strand of transparency}
While some transparency needs are addressed by XAI methods, others are met by structured documentation aimed at stakeholders not directly involved in system development,
\eg{} to help understand a system's overall design, limitations, and intended use.
Documentation frameworks exist that aim at a life cycle-scale solution \citep{chmielinski2024clear} or that support transparency for specific components of the ML pipeline: datasheets document datasets \citep{gebru2021datasheets}, model cards document trained models \citep{mitchell2019modelcards}, and care labels summarise resource usage \citep{fischer2023waschzettel}. 
On the one hand, a \stob{} itself serves as a piece of evolving documentation, falling in line with the works just mentioned but with a different focus. On the other hand, as the \stob{} aims to remain useful throughout the \bb{'s} life cycle, it itself will require accompanying documentation.



\label{ssec:CE} 

\section{Adopting Constructive Empiricism to Explainable AI}
\label{sec:CEXAI}

In this section, we translate the central notions of CE into requirements for explanatory artefacts in XAI. This translation of requirements will serve as the conceptual basis for the \textit{Scientific Theory of a \bb{}} (\stob{}), introduced in \Cref{sec:def}.
CE introduced by \citeauthor{bvf1980constructiveempiricism}, is a philosophy of science that provides a definition of a scientific theory and a description of its relation to the world and to the scientist using it.
A key feature of CE is that it does not require a theory to be literally true about the world, because the theory may include highly abstract elements or unobservable properties and processes whose correspondence with reality we cannot ascertain. Rather, a theory must be \textit{empirically adequate} and pragmatically useful for scientific practice.

We will discuss \textit{observational} and \textit{relational substructures} as the two necessary constituents of a theory, and the notions of \textit{empirical adequacy}, \textit{theory acceptance}, and \textit{pragmatic virtues}.
For the reader's convenience, \Cref{tab:operationalization} aligns the central concepts of CE with their counterparts in our setting and already anticipates the design obligations that are defined formally in \Cref{sec:def} and instantiated as a process in \Cref{sec:proc_poc}. It can already serve as a high-level guide for first-time readers, indicating the direction of subsequent sections, and it may also be referenced as a compact summary once the details have been introduced.

\begin{table}[t]
\centering
\begin{tabular}{p{0.14\textwidth} p{0.28\textwidth} p{0.50\textwidth}}
\toprule
\textbf{CE concept} & \textbf{CE definition} & \textbf{Constructing a Scientific Theory of a Black-Box} \\
\midrule 
%
%
\makecell[l]{\\Observational\\ substructures} & The measurable states and outcomes accessible to science. & Define the observation space, including inputs, targets, possible auxiliary measurements derived from the black box and respective extraction methods. Make explicit what counts as an observation and how its quality is ensured. \\ 
\makecell[l]{\\Relational\\ substructures} & The rules or relations acting on observations. & Specify the hypothesis class of the surrogate and the structural constraints that make its input–output mapping interpretable. \\ 
\makecell[l]{\\Empirical\\ adequacy} & All observable phenomena must be representable and consistent with the theory. & Ensure that the surrogate reproduces the black box on all available observations. Adequacy is binary and must be restored through updates whenever inconsistencies appear. \\ 
\makecell[l]{\\Theory\\ acceptance} & Belief in empirical adequacy and commitment to its continued use. & Ground belief in demonstrable agreement with all observations and define a clear update policy to preserve adequacy when new data arise. \\ 
\makecell[l]{\\Pragmatic\\ virtues} & Criteria to choose between empirically adequate theories on grounds of usefulness. & Identify and justify pragmatic criteria shaping the surrogate, including user-centered design choices (e.g., compactness, coherence, comprehensibility), auxiliary observables added for pragmatic reasons, and documentation with diagnostic measures that record adequacy, adaptation, and design evolution. \\ \\
\bottomrule
\end{tabular}
\caption{Conceptual alignment between CE and the design obligations of a \stob{}. It summarizes how each CE concept informs both structure and process.}
\label{tab:operationalization}
\end{table}

\subsection{Observational and Relational Substructures}
In CE's semantic view, a theory is presented as a family of structures. It distinguishes \textit{observational} from \textit{relational} substructures.
Observational substructures represent the measurable states and outcomes accessible through instruments. Measurements may be erroneous and such observations may be rejected. 
\textit{Relational substructures} in turn capture the rules or relations connecting those observations (think of variables and their composition to a formula).
Together, they provide both the descriptive surface of a theory and the framework that supports explanation. 

\textit{{Observational substructures}:}
For a \stob{}, this requires making explicit what counts as an observation: the input variables, the target outputs, and the procedures by which they are measured. As we will later discuss, an observation must include, but need not be limited to, the inputs used by the black-box (\Cref{ssec:prag_virt}).

\textit{{Relational substructures}:}
The semantic view of CE aligns with the notion of \textit{surrogate models}: a hypothesis class forms a family of admissible structures, and each instantiated surrogate corresponds to a concrete structure within that family.
For a \stob{}, this requires specifying the surrogate’s hypothesis class and the constraints imposed on it so that the mapping from inputs to outputs can be traced along an interpretable structure. 
We use \textit{traceability} to refer to the ability to follow how observations are represented and processed by the surrogate in an interpretable way and to link specific surrogate components back to the observations that constrain them.

\subsection{Empirical adequacy}
According to CE, the central aim of a scientific theory is to be \textit{empirically adequate}, that is, to be consistent with \textit{all available} observations.
A theory is empirically adequate if all observable phenomena can be represented within one of its observational substructures. 
This requires the theoretical ability to represent any possible measurement outcome as well as achieving internal consistency with all actually observed data. 
Importantly, CE does not claim that theories describe unobservable truths about the world; it is sufficient that they consistently account for all observed behaviour. 
No theory that conflicts with accepted observations can be advocated as correct. But there is no claim that the relational substructures represent the true reasoning that is being executed by the \bb{} or the hidden rules that  really govern its behaviour. 

Transferring this to surrogate models in XAI, the surrogate must reproduce the \bb{} on all currently available observations, but it needs not give guarantees for unobserved \bb{} behaviour.
This is stricter than typical surrogate modelling in XAI, where approximation error on observed data is tolerated, and looser than standard ML practice, where performance on a held-out test set is used as a proxy for generalisation to an unknown data distribution.
In the context of a \stob{}, the \bb{} represents a fixed target function, and the task is not to generalise beyond it but to reproduce its behaviour on all available observations. As the next subsection makes more explicit, this adequacy is an important building block to provide justification for trust in the surrogate.

\subsection{Theory acceptance}

For a scientist to \textit{accept} a theory means (a) to believe that it is empirically adequate, and (b) to commit to using it. 
Committing entails a willingness to work with the theory in the expectation that it will be applicable to future observations, and to adapt it if inconsistencies arise. 
In \vF{}’s view, theory building is an iterative process 
informed by observations and testing.

While empirical adequacy is concerned with the relation between the theory and the world, acceptance pertains to the relation between the theory and the scientist. We see the acceptance of a theory in CE as analogous to 
accepting a model or an explanation.
Accordingly, a \stob{} needs to give grounds to justify its acceptance, \ie{} the belief in its empirical adequacy and the decision to commit to its use.
Belief in adequacy is grounded in the surrogate \textit{demonstrably agreeing} with all available observations; the traceability requirement of the hypothesis class ensures that this agreement can be checked by linking surrogate behaviour back to specific observations.
Commitment extends this stance into the future by requiring to maintain the surrogate's adequacy: if new observations reveal inconsistencies, the surrogate must be adapted to restore adequacy and the adaptation policy must be outlined comprehensibly.
Thus, under CE, grounds for trusting the surrogate shift from one-off predictive performance on unseen data to transparent and continuous process-based guarantees.

\subsection{Pragmatic virtues}
\label{ssec:prag_virt}
Since multiple empirically adequate theories can coexist, CE recognises additional criteria for preference. 
If faced with two empirically adequate theories, it is rational to prefer one over the other solely on the basis of \textit{usefulness}, whatever this may entail for the scientist. 
Properties that are determined to be relevant to this end are subsumed under the notion of \textit{pragmatic virtues}. This may for example include choice of language, simplicity constraints, or other features that aid scientific inquiry.
Pragmatic virtues do not concern the truth of the theory but only its practical value to the scientist.

In the context of a \stob{}, two broader categories of virtues are particularly relevant:

\begin{enumerate}
    \item[(i)] \textit{User-centred criteria.} \quad
    With the goal of being useful for explanation, stakeholder-centred criteria enter already at the design stage, where several choices must be made. 
    For example, a general consensus exists regarding which broader hypothesis classes are considered inherently interpretable~\citep{molnar2020interpretable,arrieta2020xai}. 
    Designers may take these as inspiration when specifying the surrogate of a \stob{}. Each hypothesis class is typically associated with established explanation formats for which further characterisations exist that describe desirable properties~\citep{nauta2023anecdotal}. 
    With the goal of later supporting explanation, the designer might therefore consider how their chosen hypothesis class could best enable such desirable properties.

    Another important aspect is the inclusion of \textit{auxiliary measures} in the observation space that quantify \bb{} behaviours beyond the raw prediction.
    The motivation for this arises from the Rashomon effect: multiple surrogates can achieve perfect fidelity to the \bb{} while relying on different internal logics \citep{breiman2001statistical,marx2020predictive}.
    Extending the observation space with auxiliary measures, such as feature-attribution information or other observations derivable from the \bb{}, constrains adequacy on a richer set of observables, reducing this multiplicity.
    Such enrichment does not reveal whether the surrogate’s reasoning coincides with the ``real reasoning'' that  underlies the behaviour of the \bb{}, but it adds observable intermediates that make the surrogate’s behaviour empirically testable at finer resolution.
    From a CE perspective, this is a pragmatic virtue: it increases empirical discipline and user confidence without claiming access to unobservable reasoning.

    \item[(ii)] \textit{Documentation.}\quad
    A \stob{} is intended to serve not only its creators but also external stakeholders such as operators, auditors, domain experts, and affected users.
    To enable independent judgment and informed use, a \stob{} should therefore be accompanied by documentation, similar to model cards and datasheets mentioned in \Cref{sec:context}.
    It has to provide a detailed account of information related to its design, as to make assumptions, intended use, and scope explicit. It should detail how adequacy of the surrogate is determined, how its update process operates, and how other pragmatic design choices affect its behaviour.
    Analogous to how standard ML practices report learning curves, model size, or stability metrics to characterise training dynamics, the documentation can include diagnostic measures of its surrogate that record the evolution over time, \eg{} how often new data triggers updates or how its structural components take shape.
    These diagnostics are not part of empirical adequacy but complement the descriptive documentation with quantitative evidence.

\end{enumerate}

Utilising this alignment of CE to XAI concepts, we introduce SToBB, a novel artefact, in the next section.


\section{Constituents and Properties of a \stob{}}
\label{sec:def}

The previous section derived a set of concrete obligations from Constructive Empiricism (CE) and explained how they translate into operational requirements for XAI components. We now bring these elements together to define a \stob{} and describe the properties that follow from its structure. 
\Cref{sec:proc_poc} will discuss and exemplify the construction of such an artefact.

\paragraph{Constituents.}
A \textit{scientific theory of a \bb{}} (\stob{}) is a structured artefact that represents the behaviour of a fixed \bb{} through an interpretable surrogate and the processes that maintain it. A \stob{} consists of the following five components:
\begin{enumerate}
    \item \textbf{\textit{Observation base}:} An extensible record of \bb{} input-output behaviour and auxiliary measures. These observations form the empirical basis that the surrogate must match to demonstrate empirical adequacy.
    \item \textbf{\textit{Hypothesis class}:} The specification of a traceable model family that operates on the observation space.
    \item \textbf{\textit{Algorithmic components}:} Procedures for \textit{constructing} an initial surrogate and for \textit{updating} it when new observations are added, ensuring that empirical adequacy is established and maintained over time.
    \item \textbf{\textit{Adequate surrogate}:} A concrete model instantiated from the hypothesis class that agrees with all records in the observation base.
    \item \textbf{\textit{Documentation}:} A description of all theoretical, functional, and operational properties and requirements, sufficient to enable a third party to audit, deploy, and fully maintain the \stob{}.

\end{enumerate}
Each component of a \stob{} is required to satisfy one or more obligations derived in \Cref{sec:CEXAI}. In brief: \textit{Empirical adequacy} requires an explicit observation base, an instantiated surrogate, and a clearly defined adequacy criterion. \textit{Acceptance as commitment} requires an update procedure that restores adequacy when new observations arrive, and a hypothesis class that constrains the form of revisions. \textit{Pragmatic virtues} require interpretable structure, diagnostics, and documentation that support transparency, auditability, and structured use. 

\subsection{Interfaces: Extracting explanations}
\label{ssec:interfaces}

We deliberately define the role of a \stob{} to be the collection and representation of explanatory information in an empirically adequate and accessible manner; it is \textit{not} its task to answer any particular question outright.
When a concrete explanatory need arises for a given stakeholder and context, the information in the \stob{} is accessed through \textit{interfaces}.
An interface is a procedure that takes the current \stob{} and a query and returns an answer tailored to that concrete application demand. It is analogous to a way of accessing a scientific theory or applying it for some specific purpose.
Thus, interfaces are not core components of a \stob{}, and the \stob{} does not need to anticipate all future interfaces.
They are important to the use of a \stob{} but not for the conceptualisation in this paper. 
The minimal requirement for the \stob{} is that the documentation is detailed enough to enable third parties to implement interfaces for their own use-cases.

As to not undermine the intent of the \stob{}, a few general constraints for interfaces seem justified:
To provide context-appropriate answers, interfaces may transform and structure information contained in the \stob{} and add background information that the explainee requires. Since interfaces may answer questions regarding previously seen as well as unseen samples, newly encountered samples can be inconsistent with the current \stob{}.
In such cases, interfaces must not directly modify the \stob{}; instead, an update must proceed via the documented update procedure.
To preempt the disagreement problem, an interface should, if possible, preserve in its output the traceability to both the underlying observations and the surrogate structure.

In line with pragmatic virtues, interfaces defined over time may be included with the \stob{}. In that case, they should themselves be described in the documentation (purpose, assumptions, and dependencies on the \stob{}), so that their purpose and behaviour are transparent for third parties. Creators of a \stob{} will likely design interfaces for their own use, which may serve as useful examples for later maintainers.

\subsection{Properties of a \stob{}}

The following properties stem directly from the definition of a \stob{},  making it particularly valuable for XAI practice by providing a maintained and auditable point of reference across the \bb{}'s life cycle.

\textit{Baseline quality guarantees.}
Because the surrogate is required to be empirically adequate, any analysis derived from it is grounded in a model that reproduces all known behaviour of the \bb{}. Interfaces may impose additional quality checks, but these build on a demonstrably consistent foundation.

\textit{Extensibility.} The usefulness of the \stob{} is not limited to only the use-cases its designers anticipated.
Explanations are obtained through interfaces defined over the surrogate and its documentation. New types of analyses can therefore be introduced without rebuilding the \stob{}, as long as they rely on the existing structures.

\textit{Accumulation of knowledge.}
As observations accumulate, the \stob{} is strengthened because adequacy must hold for an increasingly rich empirical base. Existing results remain available, and new analyses can reuse previously established structure, which is especially valuable in time-sensitive settings.

\textit{Traceability.}
Every structural element of the surrogate can be linked back to the observations and update steps that shaped it. Explanations produced through transparent interfaces can likewise be traced to specific components of the \stob{}, forming a continuous chain of reference to underlying data.

\textit{Common point of reference.}
Because all interfaces draw from the same surrogate and documentation, their results are comparable and interpretable in a shared context. If discrepancies arise between interface outputs, the transparency of the \stob{} provides a principled basis for diagnosing and resolving them.

These properties collectively illustrate how a \stob{} offers a coherent foundation for explanation with an emphasis on usefulness throughout the life cycle of the \bb{}. They integrate the obligations of CE: empirical adequacy as the minimal requirement, acceptance through transparency and a commitment to continued adaptation, and pragmatic virtues as criteria that guide design. 
The next section describes and exemplifies the practical process by which a \stob{} is constructed and maintained.


\section{Process and Proof of Concept: Constructing a \stob{} via \cobotlong{}}
\label{sec:proc_poc}

\Cref{sec:def} defined a \stob{} through five components. We now describe a process by which a researcher may construct this artefact in practice and how it is maintained over the life cycle of a \bb{}.
To make the process concrete, we exemplify each component immediately in a proof-of-concept for a small neural network trained on a tabular task.
We introduce the \cobot{} algorithm and use it as the algorithmic component to build a \stob{}. Note that the formal description of \cobot{} is in \Cref{sec:appCoBoT}, this section provides a high-level description as to not distract from the overall proof-of-concept.

The subsections are organised around the \stob{} components \textit{observation base}, \textit{hypothesis class}, \textit{algorithmic components}, and the resulting \textit{adequate surrogate}.
In each subsection, we first state what must be specified or created, and then illustrate how this is realised in the running example.
Documentation aspects are discussed within the respective component subsections.
In addition, we discuss operational requirements, as well as the optional diagnostic information and interfaces separately at the end of the section. We collect a list of potential questions the documentation should answer in \Cref{sec:appDocform}.

Our proof-of-concept builds around a \bb{} neural network trained on the three class version of the Abalone dataset \citep{abalone_1}.
The network has a feed-forward architecture with four hidden layers, 32 neurons each, and ReLU activation functions.
The final \bb{} validation accuracy is 0.64.
The \stob{}-artefact for our proof-of-concept comprises the information given after ``\textbf{Example.}'' in each subsection, together with supplementary material in \Cref{sec:appCoBoT}, the documented code base
\footnote{Code available at 
\href{\repoURL}{\repoName}
} 
and the therein contained answers to questions in \Cref{sec:appDocform}

\subsection{Observation Base}
\label{ssec:obsbase}

The first step is to set up the \textit{observation base}, \ie{} the structures that will store observations as they are collected over the \bb{}'s life cycle.
To implement this storage, one must specify the observation \textit{space}, \ie{} the relevant \bb{} input- and output variables and, where applicable, auxiliary measures derived from the \bb{}.

The documentation should explain why each variable or measure is included, how it is obtained, and what quality criteria apply.
This enables third parties to assess whether the \stob{} rests on a relevant and reliable empirical basis.
As the \bb{} is used, new observations are recorded in the observation base and must be incorporated into the \stob{}, expanding its empirical base.

\paragraph{Example.}
The observation space consists of the input of the \bb{} (a 7 dimensional real-valued feature vector) and the corresponding class label predicted by the \bb{}.
Further, each sample is associated with attribution scores computed with LIME \citep{ribeiro2016lime}.
The attribution scores act as auxiliary observables used to constrain the surrogate to subspaces marked as important to the \bb{} decision.
The scores are obtained using the official LIME implementation\footnote{Version 0.2.0.1 from \href{https://pypi.org/project/lime/}{pypi.org/project/lime}}, using default parameters of an exponential kernel with a bandwidth of 1.93 and sampling 5000 perturbations.
An observation is rejected if all attribution scores are smaller than zero.

The observations are stored inside data fields of a ``\cobot{} class'' object (see \Cref{ssec:distop}).

\subsection{Hypothesis Class}
\label{ssec:hypclass}

The researcher needs to specify the hypothesis class of surrogate models, which determines  the structures that operate on the observations.
The documentation should describe how a surrogate processes an observation step by step, making the trace logic explicit, and specify conditions under which an observation is \textit{not} covered.
It should also record how user-centred criteria or other pragmatic virtues informed the design of the class and how these choices constrain its expressivity.


\paragraph{Example.}
As the \bb{} operates on tabular data, we choose a rule based hypothesis class, which is a popular format for explanation \citep{lakkaraju2016decisionsets}.
Rules are represented as axis-aligned bounding boxes defined over subspaces of the input domain.

Formally, let $d \in \mathbb{N}$ be the input dimensionality and let $[1..d] := \{1,\dots,d\}$.
We define the hypothesis class as a \textit{set of boxsystems}
\[
\sobs \;=\; \bigl\{\, (I, B) \;\big|\; I \subseteq [1..d],\; B = \{\bobox_1,\dots,\bobox_m\} \,\bigr\},
\]
where each pair $(I,B)$ is called a \textit{boxsystem}.
Here, $I$ denotes a fixed subspace of input dimensions and
$B$ is a finite set of \textit{non-overlapping, axis-aligned, fully bounded boxes} $\bobox$ defined on that subspace.
Each $\bobox \in B$ is associated with a class label $c \in \outputspace$.

The subspaces $I$ are derived from a
local explainer $\localexplainer$ via an indicator function $\binarization$.
Given an input sample $\sample$, $\localexplainer$ yields an attribution vector $a \in \mathbb{R}^d$, and
\[
I \;=\; \binarization\bigl(a\bigr) \subseteq [1..d]
\]
denotes the set of dimensions deemed relevant for the corresponding \bb{} prediction.
For each distinct feature set $I$ encountered in the observation base, exactly one corresponding boxsystem $(I,B)$ is maintained.

All observations whose local explanations induce the same subspace $I$ are associated with the corresponding boxsystem and constrain its boxes.
By construction, each observation is associated with exactly one subspace $I$ and, within the corresponding boxsystem, with at most one box $\bobox$.


\textit{Tracing logic of $\sobs{}$.}
Given a sample $\sample$, let
\[
I_\sample \;=\; \binarization\bigl(\localexplainer(\sample)\bigr)
\quad\text{and}\quad
c_\sample \;=\; \blbox(\sample).
\]
Tracing succeeds if there exists a boxsystem $(I_\sample, B) \in \sobs$ and a box $\bobox \in B$ such that
\[
\sample \in \bobox
\quad\text{and}\quad
\Call{Label}{\bobox} = c_\sample.
\]
If no such box exists, tracing fails and an update is triggered (see \textit{update mechanism} \Cref{ssec:algcomp}).

\textit{User-centric considerations.}
The size of $\itemset$ directly dictates the length or complexity of each rule, a factor that plays an important role in the user-friendliness of rules \citep{nauta2023anecdotal,ribeiro2018anchors}.
To control this, we introduce a parameter $k \in \mathbb{N}$ that limits the number of selected dimensions.

For a given $k$, we define a constrained indicator function $\bink$ such that, for an attribution vector  produced by $\Phi$,
\[
\bink(a) \;\subseteq\; \{\, i \in [1..d] \mid a_i > 0 \,\},
\qquad
|\bink(a)| \le k,
\]
where $\bink(a)$ contains the indices of the \textit{highest-scoring positive} components of $a$.

While this does limit the expressive power of the hypothesis class, we describe in the following subsection a principled way by which the algorithm chooses the smallest value for $k$ that allows empirical adequacy, automating the trade-off between adequacy and user-friendliness.


\subsection{Algorithmic Components}
\label{ssec:algcomp}

The algorithmic components define how a surrogate is \textit{constructed} from a set of observations and how it is \textit{updated} when adequacy fails.
The description must make the procedure reproducible and should specify whether the algorithm can represent the full hypothesis class.
The update procedure has to be described in equal detail, stating under what conditions updates are triggered, what operations may apply during updates, and how updates restore empirical adequacy.

\paragraph{Example.}
For the sake of clarity and simplicity, we focus on describing the general properties and guarantees of the algorithm. For a full technical description of \cobot{} the reader is referred to \Cref{sec:appAlg}.

\textit{Construction}: \cobot{} incrementally builds the box system, refining it when new data points are added.
The algorithm ensures that adequacy is always preserved while optimising for complexity.
We initialise \cobot{} with $k=3$ and an empty set $\sobs=\emptyset$.
Processing the observations one at a time, \cobot{} incrementally adds boxsystems to $\sobs$.
For each boxsystem the algorithm guarantees that
1) none of the boxes overlap,
2) each processed observation is unambiguously associated with one box, and
3) all samples associated with one box have the same label.
The $k=3$ dimensions used are the three dimensions with the highest attribution scores greater than zero.

\textit{Update Mechanism}: Revision is conducted in case inconsistencies arise. These can occur in three ways:
\begin{enumerate}
    \item When processing the local explanation:\quad The local explanation indicates a subspace that has not been encountered before. For this subspace, a boxsystem is created, within which a single box, containing the new observation, is placed.
    \item When processing the feature values:\quad A boxysstem for the subspace exists, but within the subspace the sample lies outside of any box. The algorithm attempts to expand the boundaries of existing boxes (of the correct label) to include the point. If an expansion is found that leads to no inconsistencies with other observations, the algorithm stops. If no possible expansion is found, the sample is placed inside a new box containing only itself.
    \item Misclassification:\quad The observation is mapped to a box with a different class label than predicted by the \bb{}. The erroneous box is then dissolved, and its contained samples are merged into existing boxes whenever possible; otherwise, new boxes are created.
\end{enumerate}

To illustrate the update mechanism, \Cref{fig:boxsystem} (\Cref{ssec:appIllustration}) shows several updates of the box-system for $\itemset = \{2,6\}$ until its current state.

\textit{Adapting maximal rule-complexity parameter $k$}:
If an incoming observation shares the same important subspace and identical features values with a previous observations, yet is associated with a different label, \cobot{} returns an error.
In this case, the two observations are indistinguishable in the selected subspace. However, since the \bb{} assigns them different values, this indicates that either
(i) the dimensionality of the subspace is too small (\ie{} the pragmatic requirement was too strict), or
(ii) the feature attribution method is not working as intended.
To address case (i), \cobot{} increases $k$ by one and rebuilds the surrogate from scratch using all stored observations, while recording its previous state and the conflicting sample for later inspection. 
Case (ii) occurs if \cobot{} fails to build an empirically adequate $\sobs$ despite using all important dimensions in an observation.
In this case, the observation space is inherently ambiguous and has to be modified externally, \eg{} by the maintainers of the \stob{} that then need to rebuilt the observation base.

\subsection{Adequate Surrogate}
\label{ssec:adsurr}
Once the observation space, hypothesis class, and algorithmic components are defined, the researcher instantiates the first surrogate. The surrogate must be empirically adequate with respect to all observations available at this stage. If adequacy fails, adjustments to the algorithm or the hypothesis class may are required before continuing. After deployment, the surrogate must be maintained: when new observations arrive that break adequacy, an update must proceed via the documented update procedure.

\paragraph{Example.}
In our example, \cobot{} instantiates the first surrogate by incrementally constructing $\sobs$ from the available observations, starting from $\sobs=\emptyset$.
Adequacy is maintained incrementally by the revision operations described before.

\subsection{Operation, distribution and maintenance}
\label{ssec:distop}

To be useful to parties other than its original creators, a \stob{} must also be documented in terms of how it is operated and distributed.
This includes specifying how the artefact is stored and versioned, and how external stakeholders can access it, \eg{} via APIs or dedicated tooling.
The documentation should outline the technical preconditions for using and updating the \stob{} (such as required software, model access, or computational resources) and the processes that govern who may perform updates and under which conditions.
Together, these provisions support structured use and maintenance of the \stob{} by third parties over the life cycle of the \bb{}.

\paragraph{Example.}
As mentioned previously, interaction with the \stob{} centres around a ``\cobot{} class'' object.
The object not only stores the observation base (\Cref{ssec:obsbase}), but also implements the algorithmic components (\Cref{ssec:algcomp}), contains the current surrogate (\Cref{ssec:adsurr}) and the record of its evolution as well as further diagnostic information (\Cref{ssec:diag} below).
The documentation of the code base contains the necessary information to load and interact with the provided object.
The code base can be found at \href{\repoURL}{\repoName}.

\FloatBarrier

\begin{figure}[t]
    \centering
    \resizebox{\textwidth}{!}{\input{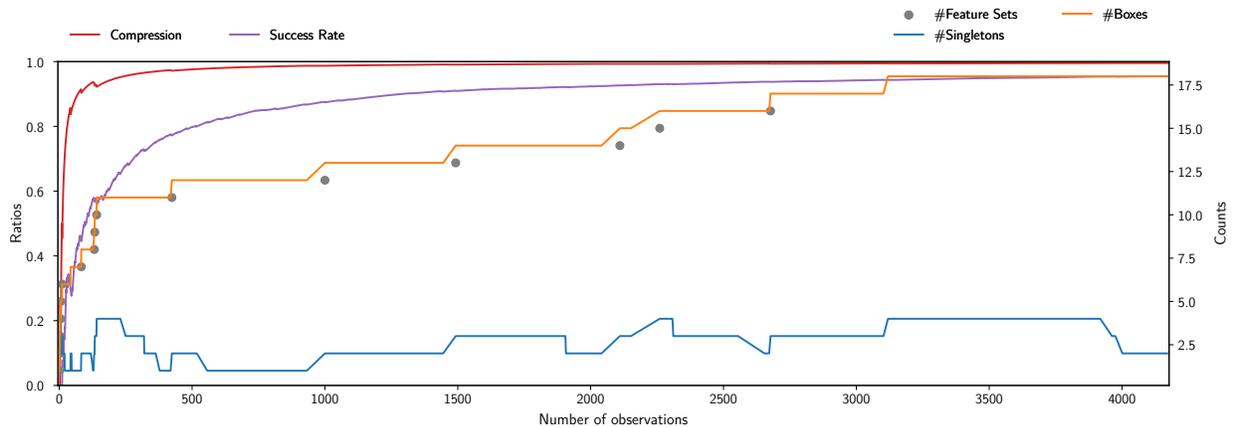}}
    \caption{Diagnostic information on the current surrogate model. Left axis (``Ratios''): Compression is defined as the gain $1-\frac{\#boxes}{\#samples}$; success rate is the fraction of samples that did \textit{not} trigger an update. Right axis (``Counts''): \#Feature Sets denotes the number of unique feature sets/distinct subspaces; \#singletons and \#boxes count the number of singleton and the total boxes, respectively. After processing all available $4\,177$ samples, the surrogate comprises less than 20 boxes. Across all observations, $187$ updates were performed.}
    \label{fig:curves}
\end{figure}
\subsection{Optional: Diagnostic Information}
\label{ssec:diag}

After the adequate surrogate has been successfully instantiated, additional diagnostic information may be collected and documented, such as coverage statistics, structural complexity measures, or the number of updates applied.
These measures do not test adequacy but complement the descriptive documentation with quantitative information.

\paragraph{Example.}
Each time \cobot{} processes an update, we track internal statistics about outcome and state of the surrogate. 
We track the \textit{total number of features sets}, the \textit{total number of boxes} across all features sets, and the \textit{total number of singletons} contained within those boxes.
\Cref{fig:curves} summarises the diagnostics over time for the current surrogate model.
The current surrogate is based on $4\,177$ observations, of which $187$ triggered updates ($4.5\%$).
Maximal subspace size remains as initialised, $k=3$. 
The boxsystems of the surrogate occupy 16 distinct subspaces and contain a cumulative sum of 20 individual boxes.
Two of those boxes are singletons, containing only a single sample.
This yields a compression gain of $1 - (20/4177)\approx 0.995$.

\subsection{Optional: Interfaces}
\label{ssec:ifaces_poc}

As discussed in \Cref{ssec:interfaces}, interfaces are not part of the \stob{} itself but may be provided and documented to better support downstream use.

\paragraph{Example.}
We include example interfaces answering three common questions in XAI that can be attributed to the leading questions \textit{Why}, \textit{How to be that}, and \textit{How}, respectively, as presented in \citet{liao2020questionbank}:
\begin{itemize}
    \item \textbf{Local interface:} \textit{Q: Why is this input given that prediction?} What features were important?
    Given an input, the interface returns a rule by extracting the box constraints relevant for that instance. This query can trigger an update on the surrogate. 
    A formal listing is given in \Cref{alg:localexp} (\Cref{ssec:appInterfaces}).
    
    \item \textbf{Contrastive interface:} \textit{Q: Why is this input given this prediction and to what values would the important features need to change to lead to a different outcome?}
    For a given input, the interface extracts the applicable box and the closest box with a different label from the same boxsystem.  This query can trigger an update on the surrogate. 
    A formal listing can be found in \Cref{alg:contrexp} (\Cref{ssec:appInterfaces}).
    
    \item \textbf{Global interface:} \textit{Q: What subspaces are considered important and how do they vary with class label and regions of the input space?}
    To answer this question, the interface first computes a 2D-UMAP projection of all input-coordinates in the observation basis.
    Each sample is then assigned to its corresponding box and  coloured according to the respective subspace (important dimensions), while class labels are encoded by marker shape.
    This provides a visual account of how the importance of different dimensions changes over the data manifold. This interface cannot trigger updates. 
    A listing of the procedure and its output are provided in \Cref{alg:glob} and \Cref{fig:umap}, respectively (\Cref{ssec:appInterfaces}).
\end{itemize}



\section{Discussion and Conclusion}
\label{sec:discussion}

We introduced the notion of a \textit{scientific theory of a black box} (\stob{}), a structured and auditable artefact that consolidates explanatory information about a fixed machine-learning model. Building on CE, we identified three obligations that guide the design of such an artefact: empirical adequacy with respect to observed behaviour, adaptability to new observations, and transparency and auditability through documentation. We translated these obligations into a general process for constructing and maintaining a \stob{} and exemplified through \cobot{} how this process can be instantiated in practice.

Following existing XAI taxonomies \citep{speith2022taxonomies}, a \stob{} can be described as a global, post-hoc artefact that centres a surrogate model. This characterisation, however, neither captures the temporal scope of the \stob{} perspective nor the particular CE-imposed obligations the surrogate has to fulfil.
More importantly, it misses what the \stob{} perspective adds beyond a surrogate description: it makes explicit a maintained representational basis that supports a clear distinction between \textit{explanatory information} and \textit{explanation}.
The explanatory information, represented by the observation base and the surrogate, can be designed and evaluated primarily with respect to empirical adequacy and traceability, independently of the interfaces that must take user-centred qualities into account.
The resulting baseline guarantees give interface designers an explicit point of reference for making principled trade-offs between information richness and the users’ need for comprehensible answers, a well-known pain point in XAI research \citep{jacovi2020towards,doshi2017towards}.
As illustrated by the \cobot{} example, a \stob{} may still be constructed with user-centred criteria in mind, but whenever there is tension, empirical adequacy takes precedence.

The aim of this paper was the conceptual development of a \stob{}. The accompanying proof-of-concept is necessarily limited in its scope:
The \cobot{} example targets one data-domain, incorporates a single auxiliary measure, and provides an online learner that covers construction and update in one procedure.
Because the \stob{} framework formulates only abstract, qualitative constraints, it leaves considerable freedom in how auxiliary measures, hypothesis classes, and algorithmic procedures are designed.
These components provide several directions for future work. Two avenues seem particularly natural:
(i) developing other hypothesis classes and algorithmic components directly derived from established interpretable model families, such as linear or logistic regression and prototype-based models \citep{molnar2020interpretable}; and
(ii) exploring existing explanation methods to assess whether and how they can be adapted into components of a \stob{}.
In our proof-of-concept, we used a local explanation method to extend the observation space with auxiliary signals of \bb{} behaviour; other methods could be adopted in a similar fashion.
Conversely, for methods such as Anchors \citep{ribeiro2018anchors} one could imagine \stob{} variants whose hypothesis class consists of sets of local rules, with algorithmic components that generalise from individual anchors and update them as new observations arrive.
Working out such adaptations requires rethinking these methods as parts of a cumulative and updateable process rather than as single-shot explainers.
The suitability of a particular observation-space design, hypothesis class, or algorithmic component will depend on the \bb{} at hand and the domain in which the AI system operates.
Over time, \stob{}s of different shapes can be explored and best practices established.

The hypothesis class of a \stob{} naturally induces a language bias that favours some types of explanations over others. In the case of \cobot{}, we used the \bb{}’s application domain to deliberately target rule-based explanations, which motivated choosing axis-aligned boxes as the surrogate structure.
As we also saw, this choice doesn't strictly limit interfaces to only produce this one type of explanation.
Interfaces are free to transform the information contained in the \stob{} as the context requires.
The work of \citeauthor{naik2020explanationspecification} may provide a useful starting point for interface design that complements the \stob{} framework particularly well. Where our paper transfers \vF{}'s Constructive Empiricism to XAI, \citet{naik2020explanationspecification} transfer \vF{}'s \textit{theory of explanation} to XAI that guides the formalisation of questions and answers.

Specifically designed to be empirically adequate and extensible, \stob{}s address use-cases that differ from those of many existing XAI methods. Their focus on accumulating information during the life cycle of a \bb{} positions them primarily as tools for AI governance.
At a minimum, a \stob{} presupposes stable access to a fixed model instance whose input-output behaviour can be logged, the ability to store and process observations in compliance with applicable legal and organisational rules, and an interpretable hypothesis class and update procedure that relevant stakeholders can inspect.
Using \stob{}s in practice will require decisions about how to regulate access, updates, and versioning, \eg{} who may read or modify a \stob{}, how documentation integrity is ensured over time, and how privacy and regulatory constraints affect what observations can be stored and which parties may query which parts of the artefact.
A \stob{} provides a structured target for such access within the bounds set by the surrounding governance regime. It cannot compensate for missing access to model behaviour, cannot override constraints on data retention and sharing, and does not in itself guarantee that all parties obtain the level of access they might ideally want.
The concrete answers to these governance questions will depend on the individual AI system and its application setting and thus lie beyond our discussion here, but they are of practical relevance to the adoption of \stob{}s and will require exploration in the future.

In this work we treated the \bb{} as fixed: once trained, its input-output behaviour does not change, \ie{} all collected observations remain valid for the lifetime of the \stob{}.
In practice, an AI system may make use of different \bb{} versions over its life cycle, \eg{} through fine-tuning or re-training.
In such cases, a \stob{} would naturally accompany a single \bb{} instance; if the model is replaced, a new \stob{} can be instantiated, potentially reusing parts of the existing observation base.
For example, in our proof-of-concept, changes to the underlying neural network would affect predictions and local explanations, but the input samples themselves remain pointers to regions of interest in the input space; this information could be used to build a new observation base.
We expect the conceptual structure of \stob{}s to be flexible enough to support such scenarios, \eg{} by integrating mechanisms in the algorithmic components, such as checks on when the observation base has become stale, or designing algorithms that allow efficient reconstruction under model change.
Working out concrete strategies for non-stationary settings is left for future work.

To conclude, by separating the persistent representational foundation of explanation from the methods that present answers in context, the \stob{} framework offers a new direction for XAI research. It complements existing XAI techniques and focuses on a coherent basis for life cycle-spanning transparency, reuse of explanatory information, and systematic external scrutiny. Future work needs to follow several directions: explore richer relational structures, extended auxiliary measures, and principled development of interfaces, as well as investigating the practical aspects of deployment and maintenance.


\bibliography{bib_sebastian}
\bibliographystyle{tmlr}

\newpage
\appendix

\section{Example Documentation Questions}
\label{sec:appDocform}
The goal of the documentation is to maximize transparency and informed use of third parties. The questions listed below may serve as a starting point:

\begin{figure*}[h]
\centering
\fbox{%
\begin{minipage}{0.95\textwidth}
\begin{multicols}{2}
\small
\noindent\textbf{Scientific Theory of a Black Box (SToBB) – Documentation Question Sheet}\\[3pt]

\textbf{\\Observations}
\begin{itemize}[leftmargin=*]
\item What variables, auxiliary measures and target outputs constitute the observation space?
\item How are auxiliary measures of the \bb{} behaviour obtained?
\item What quality criteria apply to these measurements?
\item How is the observation base stored?
\end{itemize}

\textbf{\\Hypothesis Class}
\begin{itemize}[leftmargin=*]
\item What interpretable model family defines the surrogate’s relational structure?
\item How does the surrogate process an observation step by step?
\item Under which conditions is an observation not covered by the surrogate?
\item Which user-centred criteria or pragmatic virtues shaped the design?
\end{itemize}

\textbf{\\Algorithmic Components}
\begin{itemize}[leftmargin=*]
\item How is the surrogate initially constructed from the observation base?
\item How does the surrogate demonstrate empirical adequacy?
\item How is the surrogate stored? 
\item What is the update policy when new observations appear?
\item Does the algorithm represent the full hypothesis class?
\item How are trade-offs or approximations documented?
\end{itemize}

\textbf{\\Operational Requirements}
\begin{itemize}[leftmargin=*]
\item What resources or runtime requirements must be met to deploy and maintain the \stob{}?
\item Does access to parts of the \stob{} have to be restricted for specific users or interfaces?
\item How are updates to the \stob{} performed?
\item How are updates to the \stob{} recorded?
\end{itemize}

\textbf{\\Diagnostics (if any)}
\begin{itemize}[leftmargin=*]
\item Which diagnostic metrics are monitored during construction and updates?
\item How frequently are diagnostics evaluated and recorded?
\item How are diagnostic records stored and made accessible?
\end{itemize}

\textbf{\\Interfaces (if available)}
\begin{itemize}[leftmargin=*]
\item What interfaces are implemented and what questions do they answer?
\item Who is the target audience of the interface?
\item What is the purpose, method, and assumption of each interface?
\item How can external parties audit or replicate the analyses?
\end{itemize}

\end{multicols}
\end{minipage}}

\end{figure*}

The \cobot{} repository \href{\repoURL}{\repoName} contains answers to these questions for our proof-of-concept.


\newpage 

\clearpage
\begingroup
\section{Extended documentation for the example \stob{}}
\label{sec:appCoBoT}
This section of the appendix contains a detailed description of the \cobot{} algorithm, more detailed diagnostic information from the concrete example and a formal description of the interface functions.
\subsection{Algorithmic Component \cobot{} --- Step-by-Step Description}
\label{sec:appAlg}
Following the brief description of \cobot{} in \Cref{ssec:algcomp}, we now give a detailed description of the
algorithm, going through \Cref{alg:cobot} step by step.

\begin{algorithm}[h]
    \caption{Construction and Update procedure of \cobotlong{} (\cobot)}
    \label{alg:cobot}
        \textbf{Input:} Black-box model $\blbox: \inputdomain \to \outputspace$ for some positive integer $d$ and finite set $\outputspace$, local explainer $\localexplainer$ for $\blbox$, a binarization function $\bink$ for local explanation and $k$, input sample $\sample$, set of boxsystems $\sobs$, set of encountered observations ~$\lobservation$ \\
    \textbf{Output}: Set of boxsystems $\sobs$
    
    \begin{algorithmic}[1]
        \State $c \gets \blbox(\sample)$\label{alg:fx} 
        \State $a \gets \localexplainer(\blbox, x, c)$\label{alg:complocalexp} 
        \State $I_\sample \gets \binarization(a)$\label{alg:featurest} \Comment{obtain set of important dimensions}
        \State $\lobservation \gets \lobservation \cup \{(\sample, c, I_\sample)\}\label{alg:updateobservations}$ 
        \State $\boxsystem_{I_\sample} \gets \boxsystem\, \textbf{ if } \exists(I, \boxsystem) \in \sobs \text{ with } I = I_\sample,\, \textbf{else}\, \text{None}$ 
    
        \If{$\boxsystem_{I_\sample} = \text{None}$}\label{alg:getitemsetboxes} \Comment{None on first encounter of itemset}
            \State $\bobox_\sample \gets \Call{CreateSingleton}{\sample, c, I_\sample}$\label{alg:initboxsystem}
            \State $\sobs \gets \sobs \cup \{(I_\sample, \{\bobox_\sample\})\}$\label{alg:inconsistentupdate} 
        \Else 
            \If{$\exists \bobox \in \boxsystem_{I_\sample} \text{ with } x \in \bobox$}\label{alg:boxexist}
                \If{$\Call{Label}{\bobox} \neq c$} \label{alg:wronglabel} 
                    \State $\boxsystem_{\lobservation_b} \gets \{\Call{CreateSingleton}{\observation, c_\observation, I_\observation} \mid (\observation, c_\observation, I_\observation) \in \lobservation \land I_\observation = I_\sample \land \observation \in \bobox\}$\label{alg:collectsingletons}
                    \State $\boxsystem'_{I_\sample} \gets \Call{Merge}{(\boxsystem_{I_\sample} \setminus \{\bobox\}) \cup \boxsystem_{\lobservation_\bobox}}$ 
                    \label{alg:mergsingletons}
                    \State $\sobs \gets \left(\sobs \setminus \{(I_\sample, \boxsystem_{I_\sample})\}\right) \cup \{(I_\sample, \boxsystem'_{I_\sample})\}$ 
                \EndIf
            \Else\label{alg:nobox} \Comment{no existing box contains sample}
                \State $\bobox_\sample \gets \Call{CreateSingleton}{\sample, c, I_\sample}$\label{alg:noboxsingleton} 
                \State $\boxsystem'_{I_\sample} \gets \Call{Merge}{\boxsystem_{I_\sample} \cup \{\bobox_\sample\}}$\label{alg:noboxmerge} 
                \State $\sobs \gets \left(\sobs \setminus \{(I_\sample, \boxsystem_{I_\sample})\}\right) \cup \{(I_\sample, B'_{I_\sample})\}$\label{alg:noboxupdate} 
            \EndIf
        \EndIf
        \State \Return $\sobs,\, \lobservation$ 
    \end{algorithmic}

\end{algorithm}



\textit{1) Initialization}$\ $ As input, \cobot{} requires a black-box model $\blbox$, an input sample $\sample$, a local explainer function $\localexplainer$, an indicator function $\bink$, a list of observations $\lobservation$, and a (possibly empty) set of boxsystems $\sobs$. The indicator function $\bink$ transforms a local explanation into a set of indices $\itemset$ by extracting the top $k$ most important feature dimensions.
In \cref{alg:fx,alg:complocalexp,alg:featurest}, \cobot{} computes the class label $c$ for the input sample $\sample$ and the local explanation $a$, from which it derives the feature set $\itemset_\sample$ of most important dimensions.
The observation triple $(\sample, c, \itemset_\sample)$ is then added to the set of observations $\lobservation$ in \cref{alg:updateobservations}.


\textit{2) Creating a new boxsystem}$\ $ \Cref{alg:getitemsetboxes} begins with the consistency check and the updating logic. It checks whether the feature set $\itemset_\sample$ is present in the set of boxsystems. If not, \cref{alg:initboxsystem} creates a singleton-box $\bobox_\sample$ around $\sample$ in the subspace determined by $\itemset_\sample$, associates it with the corresponding class label, and adds the tuple $(\itemset_\sample, \{\bobox_\sample\})$ to $\sobs$. \cobot{} then returns the updated set of boxsystems $\sobs$ and observations~$\lobservation$.

\textit{3) Resolving inconsistency}$\ $ In case $\itemset_\sample$ had been encountered before, \cobot{} proceeds in \cref{alg:boxexist} and checks if there exists a box $\bobox$ in the corresponding boxsystem $\boxsystem_{\itemset_\sample}$ that covers $\sample$. If so, \cref{alg:wronglabel} checks if the box's label matches the \bb{} output $c$. If the labels match, the algorithm returns. Otherwise, the boxsystem is updated:
\Cref{alg:collectsingletons} selects all samples from the current list of observations that lie within the affected box and places each into its own singleton box. In the next step (\cref{alg:mergsingletons}), the inconsistent box $\bobox$ is removed from $\boxsystem_{\itemset_\sample}$ and all singleton-boxes in $\boxsystem_{\lobservation_b}$ are attempted to be \textit{merged with} the remaining boxes in $\boxsystem_{\itemset_\sample}$.
To this end, we use an adaptation of the algorithm presented in \citet{stadtländer2024wconv}. Given a set of boxes, the algorithm greedily selects the closest two boxes having the same class, and attempts to join them. If a join leads to an overlap with any other box, the join operation is reverted. This process continues until no further consistent mergers are possible. While the original algorithm by \citeauthor{stadtländer2024wconv} was developed for binary classifications, we adapt it  mutatis mutandis to multi-class problems.
The algorithm has several desirable properties for our scenario:
\begin{itemize}
    \item[i] It naturally guarantees that \textit{all and only indicated} dimensions in $\itemset_\sample$ are used. 
    This guarantees full control over rule complexity.
    \item[ii] The algorithm guarantees to return a boxsystem that is \textit{fully} consistent with the training data, thereby fulfilling empirical adequacy. 
    \item[iii] Boxes are unambiguous: None of the computed boxes overlap, meaning that 1) each sample induces the construction of at most one box and 2) no sample can fall into two boxes 
    at the same time. 
    \item[iv] All boxes are bounded on all sides and do not extent beyond values observed in the input data. Thus, each box is strictly tied to observations.
\end{itemize}

If boxes from two different boxsystems, say $\boxsystem_{\itemset_1}$ and $\boxsystem_{\itemset_2}$, overlap, this is of no concern, because $\localexplainer$ indicated that $\blbox$ used different sets of features for its decision.
Continuing on \cref{alg:nobox}, the set of boxsystems is updated with the adapted boxsystem and \cobot{} returns.

\textit{4) Expanding coverage}$\ $  The final \textit{else} case is entered if a boxsystem does exist for $\itemset_\sample$, but none of the boxes in $\boxsystem_{\itemset_\sample}$ cover the sample. In that case \cref{alg:noboxsingleton} creates a new singleton box around $\sample$ and \cref{alg:noboxmerge} performs the merge operation described above in an attempt to extend any existing consistent box by the singleton-box. The set of boxsystems is updated and \cobot{} returns.




\subsection{Further Diagnostic Information}
\label{ssec:appIllustration}

This further complements the diagnostic information shown in the curves in \Cref{fig:curves}, \Cref{ssec:diag}. 


\paragraph{Increments of maximal subspace size} \cobot{} was initialized with $k=3$, this has not been increased.

\paragraph{Evolution of a box-system} Despite the current subspace size being $k=3$, the surrogate uses subspaces with less than $3$ dimensions.
\Cref{fig:boxsystem} shows the subspace of the boxsystem for feature set $\itemset=\{2,6\}$ at different points in time. Two classes are present, indicated by the two different box colors. Samples of individual observations are visualized as dots. The scatterplot indicates that the features are linearly correlated within the subspace. Of the 187 total updates to the surrogate \cobot{} performed, 27 were performed on this boxsystem.

\begin{figure}[h]
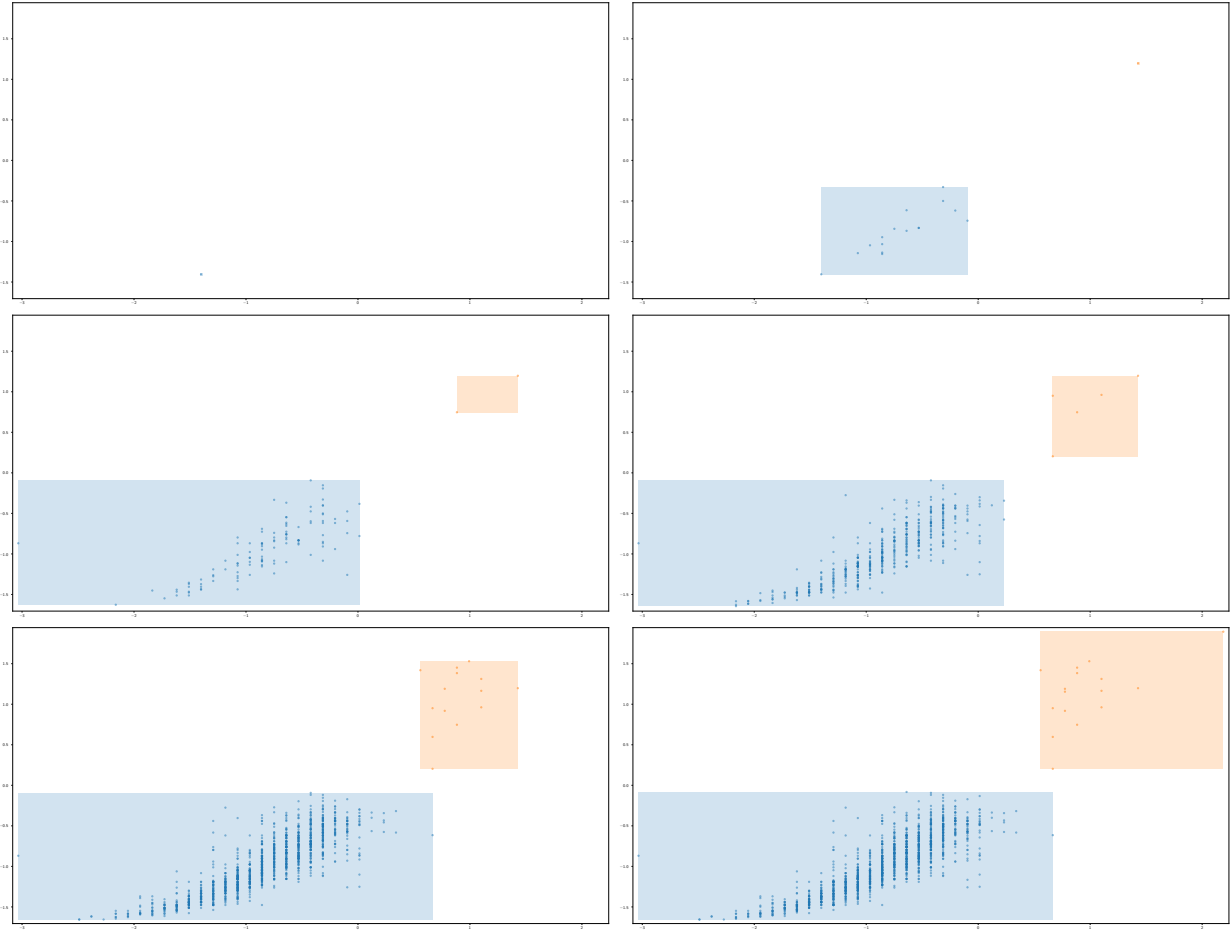

 \centering 
    \begin{adjustbox}{width=\linewidth}
        \begin{tabular}{c c}
    \input{figures/TMLR/boxsystem/0.pgf} & \input{figures/TMLR/boxsystem/6.pgf} \\[1ex]
    \input{figures/TMLR/boxsystem/15.pgf} & \input{figures/TMLR/boxsystem/18.pgf} \\[1ex]
    \input{figures/TMLR/boxsystem/24.pgf} & \input{figures/TMLR/boxsystem/26.pgf} \\[1ex]
\end{tabular}

    \end{adjustbox}
 \caption{Evolution of the axis-aligned bounding boxes contained in the boxsystem for subapace $\itemset=\{2,6\}$. Each individual image shows the same boxystem after different updates. Updates number [1, 7, 16, 19, 25, 27 (final)]. Colors encode classes.}
 \label{fig:boxsystem}
\end{figure}

\clearpage
\subsection{Interfaces}
\label{ssec:appInterfaces}

We provide interfaces addressing categories of questions listed in the XAI question bank~\citep{liao2020questionbank}.
By design, the surrogate's hypothesis class naturally supports extracting rule based explanations. \Cref{alg:localexp} and \Cref{alg:contrexp} extract a local and a contrastive explanation, respectively. \Cref{alg:glob} gives a global view. 

\Cref{alg:localexp} answers the question ``Given $\sample$, why did $\blbox$ predict $c$?'' by returning the applicable box.
By instead returning the supporting samples of the relevant box, a different interface could give an ``explanation by example'', answering the question ``What kind of instance gets the same prediction'' . 
\Cref{alg:contrexp} answers the question ``Given $\sample$, why did $\blbox$ predict $c$ and what other class is the sample most similar to?'', supplying in its output also a neighboring box of a different class.
It could be adapted to give a targeted answer ``Why $c$ and not $c'\ $?'' by restricting the search in \href{line:3search}{line 9} to $\bobox$ associated with $c'$. In case the request from any of the interfaces leads to a failure condition (\textit{itemset unknown, not covered, wrong label}), \cobot{} automatically updates the surrogate.
\begin{figure}[h]
\input{appendix/interfaces_local_contr}
\end{figure}

To obtain a global view on $\sobs$, we define \Cref{alg:glob}. It computes a 2d UMAP embedding of all observations, color-coded by the feature set they were mapped to and the marker shape indicating the class label predicted by $\blbox$.  \Cref{fig:umap} shows the interface outputs for the current set of the \stob{}.
The color imbalance highlights that the feature sets have vastly different support.
Feature set $\{2,6\}$ (light pink) occupies most of the right. Despite the comparatively large support of the feature set, the number of updates it triggered was below 15\% of all updates (27/187).
The left side of the plot is occupied by more feature sets, brown, blue and green.
Purple, lavender, cyan and orange have very little support and are confined to very small areas.
Colours form perceptibly connected regions.
Conspicuously, many feature sets seem to use only a single type of marker, indicating that classes are separable not only by value but by feature alone. More detailed statistics could be delivered by another interface.


\begin{algorithm}
    \caption{Global Visualization}
    \label{alg:glob}
    \textbf{Input:} Set of observations $\lobservation = \{(\sample, c_\sample, I_\sample)\}$, a 2D projection operator $\text{UMAP}(\cdot)$\\
    \textbf{Output}: 2D visualization of projected samples colored by $I_\sample$ and shaped by $c_\sample$
    \begin{algorithmic}[1]
        \State $X \gets \{\sample \mid (\sample, c_\sample, I_\sample) \in \lobservation\}$
        \label{alg:glob-collectX}
        \State $(u_\sample, v_\sample)_{\sample \in X} \gets \text{UMAP}(X)$
        \label{alg:glob-umap}
        \State Visualization $\gets$ \Call{EmptyPlot}{}
        \For{each $(\sample, c_\sample, I_\sample) \in \lobservation$}
            \State $\text{color}_\sample \gets $\Call{SubspaceToColor}{$I_\sample$}\Comment{color encodes important dimensions}
            \State $\text{marker}_\sample \gets $\Call{ClassToMarker}{$c_\sample$} \Comment{marker encodes class label}
            \State Visualization.\Call{AddScatter}{$(u_x, v_x)$, color$_{\sample}$, marker$_{\sample}$}
        \EndFor

        \State \Return Visualization
    \end{algorithmic}
\end{algorithm}

\begin{figure}[]
    \centering
    \resizebox{\textwidth}{!}{
        \includegraphics{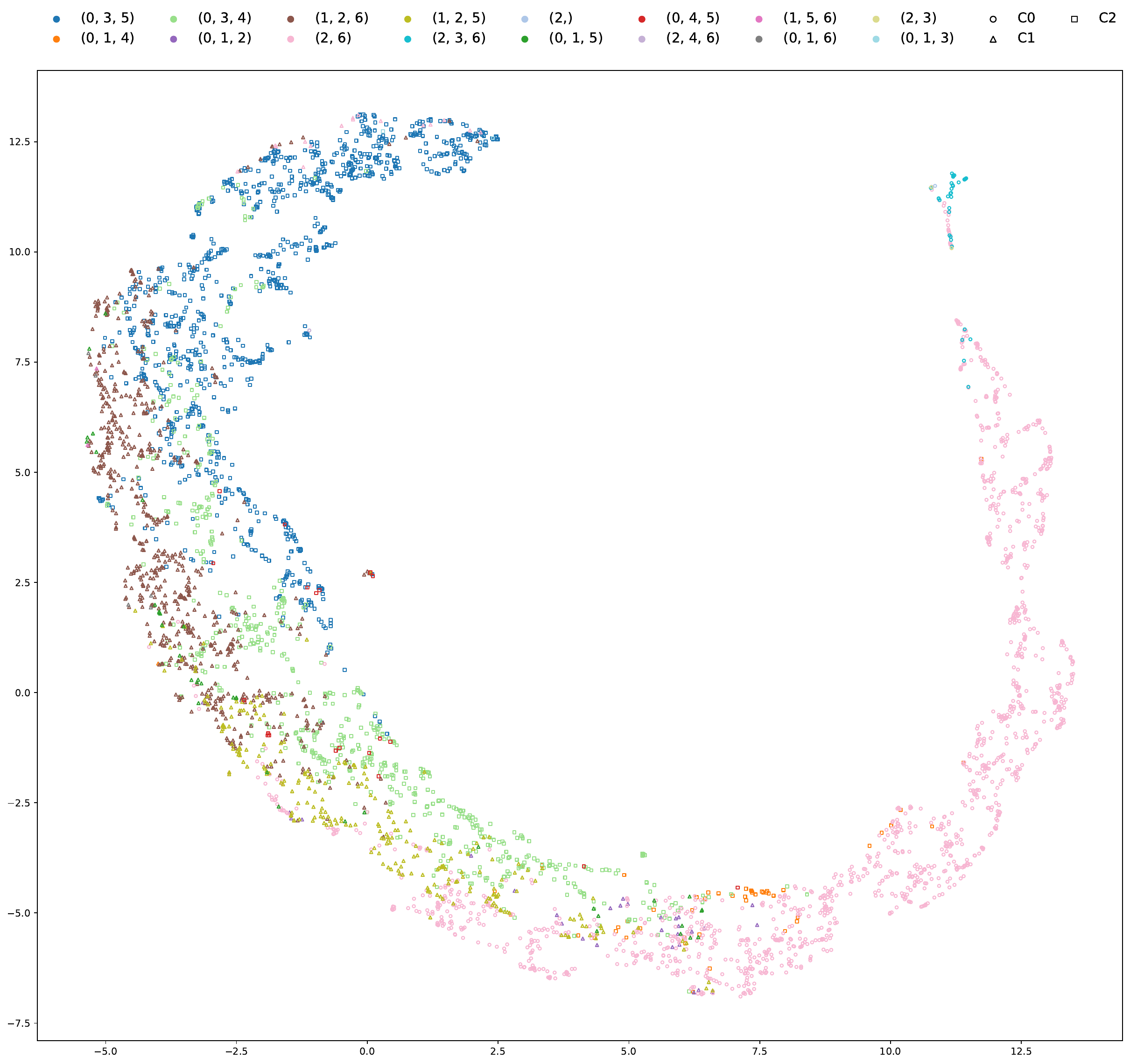} 
    }
    \caption{
    2d UMAP embedding of the 4\,177 observed samples in the \cobot{}-\stob{}. 
    Colors indicate feature sets from $\binthree$, marker style indicates class label.
    }
    \label{fig:umap}
\end{figure}

\endgroup


\end{document}